\documentclass[runningheads]{llncs}
\usepackage{graphicx}
\usepackage{comment}
\usepackage{amsmath,amssymb}
\usepackage{color}
\usepackage{url}
\usepackage{booktabs, multirow, array, makecell, caption}
\usepackage{verbatimbox}

\begin{document}

\pagestyle{headings}
\mainmatter
\def\ECCVSubNumber{4522}

\title{Polarimetric Multi-View Inverse Rendering}

\titlerunning{Polarimetric MVIR}

\author{Jinyu Zhao \and Yusuke Monno \and Masatoshi Okutomi}

\authorrunning{J. Zhao et al.}

\institute{Tokyo Institute of Technology, Tokyo, Japan}

\maketitle

\begin{abstract}
A polarization camera has great potential for 3D reconstruction since the angle of polarization~(AoP) of reflected light is related to an object's surface normal. In this paper, we propose a novel 3D reconstruction method called Polarimetric Multi-View Inverse Rendering~(Polarimetric MVIR) that effectively exploits geometric, photometric, and polarimetric cues extracted from input multi-view color polarization images. We first estimate camera poses and an initial 3D model by geometric reconstruction with a standard structure-from-motion and multi-view stereo pipeline. We then refine the initial model by optimizing photometric rendering errors and polarimetric errors using multi-view RGB and AoP images, where we propose a novel polarimetric cost function that enables us to effectively constrain each estimated surface vertex's normal while considering four possible ambiguous azimuth angles revealed from the AoP measurement. Experimental results using both synthetic and real data demonstrate that our Polarimetric MVIR can reconstruct a detailed 3D shape without assuming a specific polarized reflection depending on the material.
\keywords{Multi-view reconstruction, inverse rendering, polarization}
\end{abstract}

\section{Introduction}

Image-based 3D reconstruction has been studied for years and can be applied to various applications, e.g. model creation~\cite{biehler20143d}, localization~\cite{cao2013graph}, segmentation~\cite{dai20183dmv}, 
and shape recognition~\cite{su2015multi}. There are two common approaches for 3D reconstruction: geometric reconstruction and photometric reconstruction. The geometric reconstruction is based on feature matching and triangulation using multi-view images. It has been well established as structure from motion~(SfM)~\cite{agarwal2009building,schonberger2016structure,wu2011high} for sparse point cloud reconstruction, often followed by dense reconstruction with multi-view stereo~(MVS)~\cite{furukawa2010towards,furukawa2009accurate,galliani2015massively}. On the other hand, the photometric reconstruction exploits shading information for each image pixel to derive dense surface normals. It has been well studied as shape from shading~\cite{barron2014shape,xiong2014shading,zhang1999shape} and photometric stereo~\cite{haefner2019variational,ikehata2014photometric,wu2010robust}.

There also exist other advanced methods combining the advantages of both approaches, e.g. multi-view photometric stereo~\cite{li2020multi,park2016robust} and multi-view inverse rendering~(MVIR)~\cite{kim2016multi}. These methods typically start with SfM and MVS for camera pose estimation and initial model reconstruction, and then refine the initial model, especially for texture-less surfaces, by utilizing shading cues.

\begin{figure}[t!]
\centering
\includegraphics[trim={0cm 0cm 0cm 0cm}, width=1.0\linewidth]{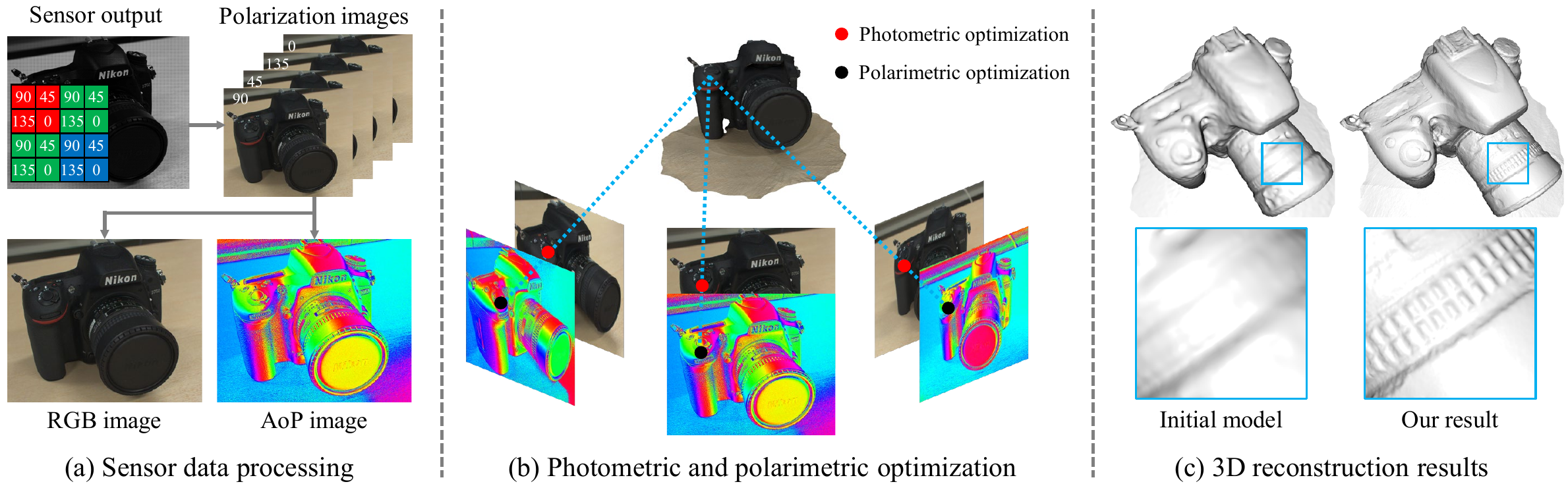}
\caption{Overview of our Polarimetric Multi-View Inverse Rendering~(Polarimetric MVIR): (a) Color polarization sensor data processing to obtain the set of RGB and angle-of-polarization~(AoP) images; (b) Using estimated camera poses and an initial model from SfM and MVS, Polarimetric MVIR optimizes photometric rendering errors and polarimetric errors by using multi-view RGB and AoP images; (c) Initial and our refined 3D model results.}
\label{fig:overall}
\end{figure}

Multi-view reconstruction using polarization images~\cite{cui2017polarimetric,yang2018polarimetric} has also received increasing attention with the development of one-shot polarization cameras using Sony IMX250 monochrome or color polarization sensor~\cite{maruyama20183}, e.g. JAI GO-5100MP-PGE~\cite{JAI} and Lucid PHX050S-Q~\cite{Lucid} cameras. The use of polarimetric information has great potential for 3D reconstruction since the angle of polarization~(AoP) of reflected light is related to the azimuth angle of the object's surface normal. One state-of-the-art method is Polarimetric MVS~\cite{cui2017polarimetric}, which propagates initial sparse depth from SfM by using AoP images obtained by a polarization camera for creating a dense depth map for each view. 
Since there are four possible azimuth angles corresponding to one AoP measurement as detailed in Section~\ref{sec:p-ambiguities}, their depth propagation relies on the disambiguation of polarimetric ambiguities using the initial depth estimate by SfM.

In this paper, inspired by the success of MVIR \cite{kim2016multi} and Polarimetric MVS~\cite{cui2017polarimetric}, we propose Polarimetric Multi-View Inverse Rendering (Polarimetric MVIR), which is a fully passive 3D reconstruction method exploiting all geometric, photometric, and polarimetric cues. We first estimate camera poses and an initial surface model based on SfM and MVS. We then refine the initial model by simultaneously using multi-view RGB and AoP images obtained from color polarization images~(see Fig.~\ref{fig:overall}) while estimating surface albedos and illuminations for each image. The key of our method is a novel global cost optimization framework for shape refinement. In addition to a standard photometric rendering term that evaluates RGB intensity errors (as in \cite{kim2016multi}), we introduce a novel polarimetric term that evaluates the difference between the azimuth angle of each estimated surface vertex's normal and four possible azimuth angles obtained from the corresponding AoP measurement. Our method takes all four possible ambiguous azimuth angles into account in the global optimization, instead of explicitly trying to solve the ambiguity as in Polarimetric MVS~\cite{cui2017polarimetric}, which makes our method more robust to noise and mis-disambiguation. Experimental results using synthetic and real data demonstrate that, compared with existing MVS methods, MVIR, and Polarimetric MVS, Polarimetric MVIR 
can reconstruct a more detailed 3D model from unconstrained input images without any prerequisites for surface materials. Two main contributions of this work are summarized as below.
\begin{itemize}
    \item We propose Polarimetric MVIR, which is the first 3D reconstruction method based on multi-view photometric and polarimetric optimization with an inverse rendering framework.
    \item We propose a novel polarimetric cost function that enables us to effectively constrain the surface normal of each vertex of the estimated surface mesh while considering the azimuth angle ambiguities as an optimization problem. 
\end{itemize}

\section{Related Work}
In the past literature, a number of methods have been proposed for the geometric 3D reconstruction~(e.g. SfM~\cite{agarwal2009building,schonberger2016structure,wu2011high} and MVS~\cite{furukawa2010towards,furukawa2009accurate,galliani2015massively}) and the photometric 3D reconstruction~(e.g. shape from shading~\cite{barron2014shape,xiong2014shading,zhang1999shape} and photometric stereo~\cite{haefner2019variational,ikehata2014photometric,wu2010robust}).
In this section, we briefly introduce the combined methods of geometric and photometric 3D reconstruction, and also polarimetric 3D reconstruction methods, which are closely related to our work. 

\noindent
{\bf Multi-view geometric-photometric 3D reconstruction:} The geometric approach is relatively robust to estimate camera poses and a sparse or dense point cloud, owing to the development of robust feature detection and matching algorithms~\cite{bay2008speeded,lowe2004distinctive}. However, it is weak in texture-less surfaces because sufficient feature correspondences cannot be obtained. In contrast, the photometric approach can recover fine details for texture-less surfaces by exploiting pixel-by-pixel shading information. However, it generally assumes a known or calibrated camera and lighting setup. Some advanced methods~\cite{maurer2016combining,wu2010fusing,wu2011high}, including multi-view photometric stereo~\cite{li2020multi,park2016robust} and MVIR~\cite{kim2016multi}, combine the two approaches to take both advantages. These methods typically estimate camera poses and an initial model based on SfM and MVS, and then refine the initial model, especially in texture-less regions, by using shading cues from multiple viewpoints. Our Polarimetric MVIR is built on MVIR~\cite{kim2016multi}, which is an uncalibrated method and jointly estimates a refined shape, surface albedos, and each image's illumination.

\noindent
{\bf Single-view shape from polarization~(SfP):} There are many SfP methods which estimate object's surface normals~\cite{atkinson2006recovery,huynh2013shape,kadambi2015polarized,miyazaki2003polarization,morel2005polarization,smith2018height,tozza2017linear} based on the physical properties that AoP and degree of polarization (DoP) of reflected light are related to the azimuth and the zenith angles of the object's surface normal, respectively.
However, existing SfP methods usually assume a specific surface material because of the material-dependent ambiguous relationship between AoP and the azimuth angle, and also the ambiguous relationship between DoP and the zenith angle. For instance, a diffuse polarization model is adopted in~\cite{atkinson2006recovery,huynh2013shape,miyazaki2003polarization,tozza2017linear}, a specular polarization model is applied in~\cite{morel2005polarization}, and dielectric material is considered in~\cite{kadambi2015polarized,smith2018height}. Some methods combine SfP with shape from shading or photometric stereo~\cite{atkinson2017polarisation,mahmoud2012direct,ngo2015shape,miyazaki2003polarization,smith2018height,zhu2019depth}, where estimated surface normals from shading information are used as cues for resolving the polarimetric ambiguity. However, these methods require a calibrated lighting setup.

\noindent
{\bf Multi-view geometric-polarimetric 3D reconstruction:} Some studies have shown that multi-view polarimetric information is valuable for surface normal estimation~\cite{atkinson2007shape,ghosh2011multiview,miyazaki2020shape,miyazaki2016surface,rahmann2001reconstruction} and also camera pose estimation~\cite{chen2018polarimetric,cui2019polarimetric}. However, existing multi-view methods typically assume a specific material, e.g. diffuse objects~\cite{atkinson2007shape,cui2019polarimetric}, specular objects~\cite{miyazaki2020shape,miyazaki2016surface,rahmann2001reconstruction} and faces~\cite{ghosh2011multiview}, to omit the polarimetric ambiguities. Recent two state-of-the-art methods, Polarimetric MVS~\cite{cui2017polarimetric} and Polarimetric SLAM~\cite{yang2018polarimetric}, consider a mixed diffuse and specular reflection model to remove the necessity of known surface materials. These methods first disambiguate the ambiguity for AoP by using initial sparse depth cues from MVS or SLAM. Each viewpoint's depth map is then densified by propagating the sparse depth, where the disambiguated AoP values are used to find iso-depth contours along which the depth can be propagated. Although dense multi-view depth maps can be generated by the depth propagation, this approach relies on correct disambiguation which is not easy in general.

\noindent
{\bf Advantages of Polarimetric MVIR:} Compared to prior studies, our method has several advantages. First, it advances MVIR~\cite{kim2016multi} by using polarimetric information while inheriting the benefits of MVIR. Second, similar to~\cite{cui2017polarimetric,yang2018polarimetric}, our method is fully passive and does not require calibrated lighting and known surface materials. Third, polarimetric ambiguities are resolved as an optimization problem in shape refinement, instead of explicitly disambiguating them beforehand as in~\cite{cui2017polarimetric,yang2018polarimetric}, which can avoid relying on the assumption that the disambiguation is correct. Finally, a fine shape can be obtained by simultaneously exploiting photometric and polarimetric cues, where multi-view AoP measurements are used for constraining each estimated surface vertex's normal, which is a more direct and natural way to exploit azimuth-angle-related AoP measurements for shape estimation.

\section{Polarimetric Ambiguities in Surface Normal Prediction}
\label{sec:p-ambiguities}

\subsection{Polarimetric calculation}
\label{subsec:calculation}
Unpolarized light becomes partially polarized after reflection by a certain object's surface. Consequently, under common unpolarized illumination, the intensity of reflected light observed by a camera equipped with a polarizer satisfies the following equation:
\begin{equation}
   I(\phi_{pol})=\frac{I_{max}+I_{min}}{2}+\frac{I_{max}-I_{min}}{2}
   {\rm cos}2(\phi_{pol}-\phi),
\end{equation}
where $I_{max}$ and $I_{min}$
are the maximum and minimum intensities, respectively, $\phi_{pol}$ is the polarizer angle, and $\phi$ is the reflected light's AoP, which indicates reflection's direction of polarization.
A polarization camera commonly observes the intensities of four polarization directions, i.e. $I_0$, $I_{45}$, $I_{90}$, and $I_{135}$. From those measurements, AoP can be calculated using the Stokes vector~\cite{stokes1851composition} as 
\begin{equation}
\label{eq:AoP}
   \phi=\frac{1}{2}{\rm tan}^{-1}\frac{s_2}{s_1},
\end{equation}
where $\phi$ is the AoP, and $s_1$ and $s_2$ are the components of the Stokes vector 
\begin{equation}
\label{eq:Stokes}
   \textbf{s}=\left[\begin{matrix}
      s_0\\s_1\\s_2\\s_3
   \end{matrix}
   \right]
   =\left[\begin{matrix}
      I_{max}+I_{min}\\(I_{max}-I_{min}){\rm cos}(2\phi)\\(I_{max}-I_{min}){\rm sin}(2\phi)\\0
   \end{matrix}
   \right]
   =\left[\begin{matrix}
      I_0+I_{90}\\I_0-I_{90}\\I_{45}-I_{135}\\0
   \end{matrix}
   \right],
\end{equation}
where $s_3=0$ because circularly polarized light is not considered in this work.

\subsection{Ambiguities}
\label{sec:ambiguities}

\begin{figure}[t!]
\centering
\includegraphics[trim={0cm 0cm 0cm 0cm}, width=1\linewidth]{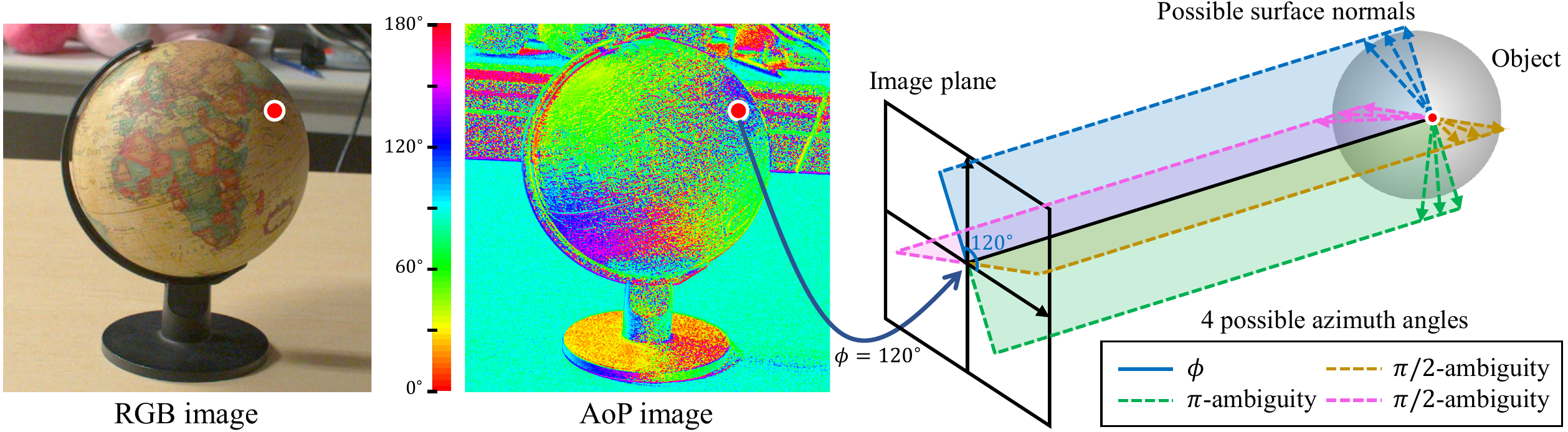}
\caption{Four possible azimuth angles ($\alpha=30^{\circ},\ 120^{\circ},\ 210^{\circ}$ and $300^{\circ}$) corresponding to an observed AoP value ($\phi=120^{\circ}$). The transparent color plane shows the possible planes on which surface normal lies. Example possible surface normals are illustrated by the color dashed arrows on the object.}
\label{fig:ambiguity}
\end{figure}

AoP of reflected light reveals
information about the surface normal according to Fresnel equations,
as depicted by Atkinson and Hancock \cite{atkinson2006recovery}.
There are two linear polarization components of the incident wave:
$s$-polarized light and $p$-polarized light
whose directions of polarization are perpendicular and parallel to
the plane of incidence consisting of incident light and surface normal, respectively. 
For a dielectric, the reflection coefficient
of $s$-polarized light is always greater than that of $p$-polarized light
while the transmission coefficient of $p$-polarized light
is always greater than that of $s$-polarized light. For a metal, the relationship are opposite.
Consequently, the polarization direction of reflected light should be perpendicular or parallel to the plane of incidence according to the relationship between 
$s$-polarized and $p$-polarized light.

In this work, we consider a mixed polarization reflection model~\cite{baek2018simultaneous,cui2017polarimetric}
which includes unpolarized
diffuse reflection, polarized specular reflection 
($s$-polarized light is stronger)
and polarized diffuse reflection 
($p$-polarized light is stronger).
In that case, the relationship between AoP and 
the azimuth angle, which is the angle between surface normal's
projection to the image plane and $x$-axis in the image coordinates,
depends on which polarized reflection's component is dominant.
In short, as illustrated in Fig.~\ref{fig:ambiguity}, there exist two kinds of ambiguities.

\noindent
{\bf $\pi$-ambiguity:}
$\pi$-ambiguity exists because the range of AoP is from 0 to $\pi$ 
while that of the azimuth angle is from 0 to $2\pi$.
AoP corresponds to the same direction
or the inverse direction of the surface normal, i.e. AoP may be equal to the
azimuth angle or have $\pi$'s difference with the azimuth angle. 

\noindent
{\bf $\pi/2$-ambiguity:}
It is difficult to decide whether polarized specular reflection
or polarized diffuse reflection dominates without 
any prerequisites for surface materials. 
AoP has $\pi/2$'s difference with the azimuth angle when polarized specular 
reflection dominates, 
while it equals the azimuth angle or has $\pi$'s difference with the azimuth angle when polarized diffuse reflection dominates.
Therefore, there exists $\pi/2$-ambiguity in addition to $\pi$-ambiguity
when determining the relationship between AoP and the azimuth angle.

As shown in Fig.~\ref{fig:ambiguity}, for the AoP value ($\phi=120^\circ$) for the pixel marked in red,
there are four possible azimuth angles (i.e. $\alpha=30^{\circ},\ 120^{\circ},\ 210^{\circ}$ and $300^{\circ}$)
as depicted by the four lines on the image plane.
The planes where the surface normal has to lie, which are represented by the four transparent color planes, are determined according to the four possible azimuth angles. The dashed arrows on the object show the examples of possible surface normals, which are constrained on the planes. In our method, the explained relationship between the AoP measurement and the possible azimuth angles is exploited to constrain the estimated surface vertex's normal.

\section{Polarimetric Multi-View Inverse Rendering}
\label{sec:pmvir}
\subsection{Color polarization sensor data processing}
\label{subsec:rawDataProcessing}

To obtain input RGB and AoP images, we use a one-shot color polarization camera consisting of the $4\!\times\!4$ regular pixel pattern~\cite{maruyama20183} as shown in Fig.~\ref{fig:overall}(a), although our method is not limited to this kind of polarization camera. For every pixel, twelve values, i.e. $3\ (R,G,B)\times4\ (I_0,I_{45},I_{90},I_{135})$, are obtained
by interpolating the raw mosaic data.
As proposed in \cite{morimatsu2020}, 
pixel values for each direction in every $2\!\times\!2$ blocks are extracted to obtain Bayer-patterned data for that direction. Then, Bayer color interpolation~\cite{kiku2016beyond} and polarization interpolation~\cite{mihoubi2018survey} are sequentially performed to obtain full-color-polarization data. As for the RGB images used for the subsequent processing, we employ unpolarized RGB component ${\bf I}_{min}$ obtained as ${\bf I}_{min} = ({\bf I}_{0}+{\bf I}_{90})(1-\rho)/2$, where $\rho$ is DoP and calculated by using the Stokes vector of Eq.~(\ref{eq:Stokes}) as $\rho = \sqrt{s_1^2+s_2^2}/{s_0}$. 
Since using $\textbf{I}_{min}$ can suppress the influence of specular reflection~\cite{atkinson2006recovery}, it is beneficial for SfM and our photometric optimization.
On the other hand, AoP values are calculated using Eq.~(\ref{eq:AoP}) and (\ref{eq:Stokes}), where the intensities of four polarization directions ($I_0$, $I_{45}$, $I_{90}$, $I_{135}$) are obtained by averaging R, G, and B values for each direction.

\begin{figure}[t!]
\centering
\includegraphics[trim={0cm 0cm 0cm 0cm}, width=1\linewidth]{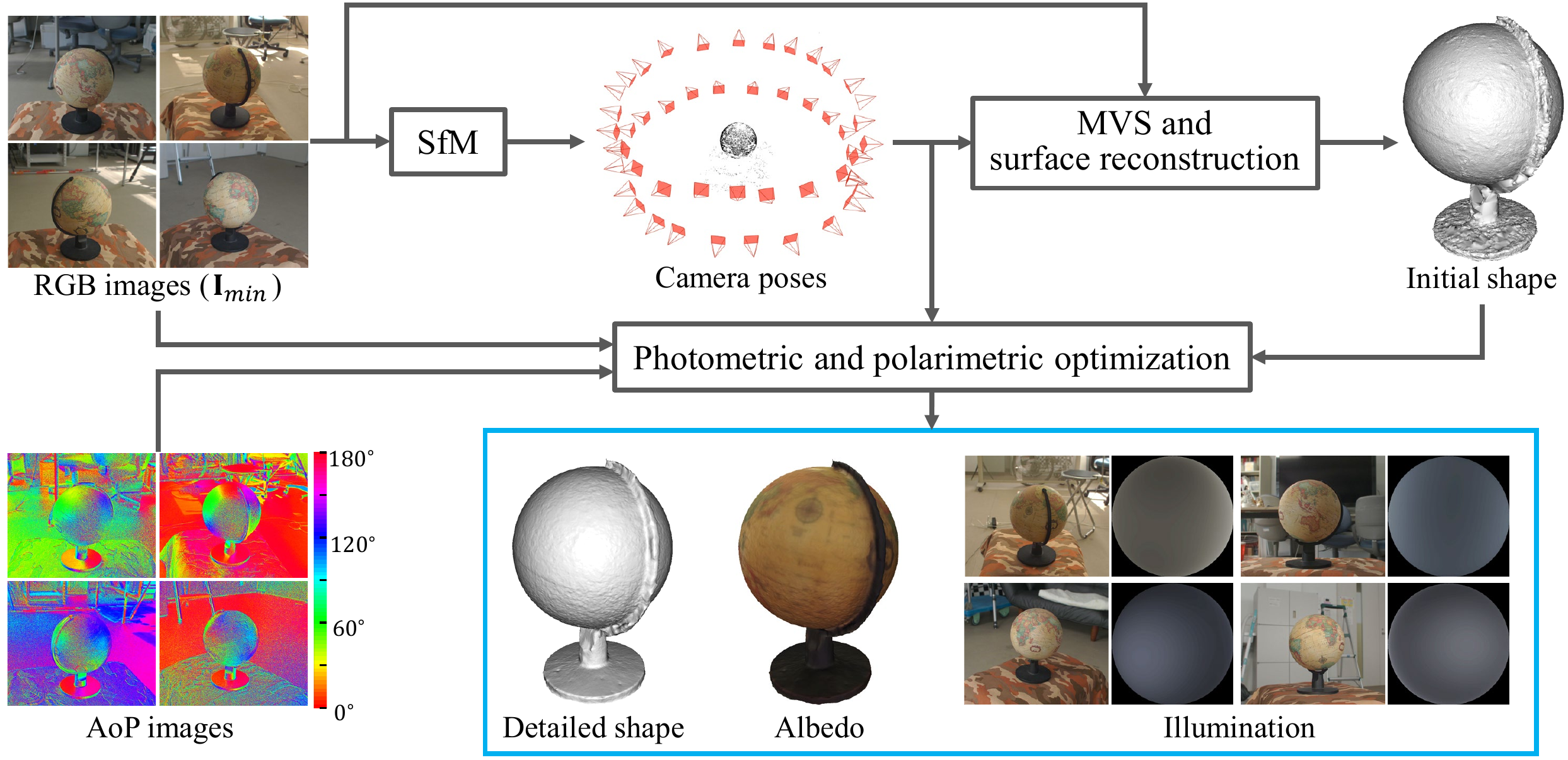}
\caption{The flowchart of our Polarimetric MVIR using multi-view RGB and AoP images.}
\label{fig:flowchart}
\label{flowchart-test}
\end{figure}

\subsection{Initial geometric reconstruction}
Figure~\ref{fig:flowchart} shows the overall flow of our Poralimetric MVIR using multi-view RGB and AoP images. It starts with initial geometric 3D reconstruction as follows. SfM is firstly performed using the RGB images to estimate camera poses. Then, MVS and surface reconstruction are applied to obtain an initial surface model which is represented by a triangular mesh. The visibility of each vertex to each camera is then checked using the algorithm in~\cite{kim2016multi}. Finally, to increase the number of vertices, the initial surface is subdivided by $\sqrt{3}$-subdivision~\cite{kobbelt20003} until the maximum pixel number in each triangular patch projected to visible cameras becomes smaller than a threshold. 

\subsection{Photometric and polarimetric optimization}
\label{sec:optimize}
The photometric and polarimetric optimization is then performed to refine the initial model while estimating each vertex's albedo and each image's illumination. The cost function is expressed as 
\begin{equation}
   \label{eq:costFunction}
   \mathop{\arg\min}\limits_{{\bf X}, {\bf K}, {\bf L}} E_{pho}({\bf X}, {\bf K}, {\bf L}) + \tau_1 E_{pol}({\bf X})
   + \tau_2 E_{gsm}({\bf X}) + \tau_3 E_{psm}({\bf X}, {\bf K}),
\end{equation}
where $E_{pho}$, $E_{pol}$, $E_{gsm}$, and $E_{psm}$ represent a photometric rendering term, a polarimetric term, a geometric smoothness term, and a photometric smoothness term, respectively. $\tau_1$, $\tau_2$, and $\tau_3$ are weights to balance each term. Similar to MVIR~\cite{kim2016multi}, the optimization parameters are defined as below:

\noindent
- ${\bf X}\in\mathbb{R}^{3\times n}$ is the 
vertex 3D coordinate, where $n$ is the total number
of vertices.

\noindent
- ${\bf K}\in\mathbb{R}^{3\times n}$ is the vertex albedo,
which is expressed in the RGB color space.

\noindent
- ${\bf L}\in\mathbb{R}^{12\times p}$ is the scene illumination matrix, where $p$
is the total number of images. Each image's illumination is represented by nine coefficients for the second-order spherical harmonics basis $(L_0,\cdots,L_8)$~\cite{ramamoorthi2001efficient,wu2011high} and three RGB color scales $(L_R,L_G,L_B)$.

\noindent
{\bf Photometric rendering term:}
We adopt the same photometric rendering term as MVIR, which is expressed as
\begin{equation}
\label{photometricRenderingTerm}
   E_{pho}({\bf X}, {\bf K}, {\bf L})=\sum_i \sum_{c\in \mathcal{V}(i)} \frac{
      ||{\bf I}_{i,c}({\bf X})-\hat{{\bf I}}_{i,c}({\bf X}, {\bf K}, {\bf L})||^2}{|\mathcal{V}(i)|},
\end{equation}
which measures the pixel-wise intensity error between observed and rendered values.
${\bf I}_{i,c}\in\mathbb{R}^3$ is the observed RGB values of the pixel in $c$-th image corresponding to $i$-th vertex's projection and $\hat{{\bf I}}_{i,c}\in \mathbb{R}^3$ is the corresponding rendered RGB values. 
$\mathcal{V}(i)$ represents the visible camera set for $i$-th vertex. The perspective projection model is used to project each vertex to each camera.
Suppose $(K_R,K_G,K_B)$ and $(L_0,\cdots,L_8,L_R,L_G,L_B)$ represent the albedo for $i$-th vertex and the illumination for $c$-th image, where the indexes $i$ and $c$ are omitted for notation simplicity.  
The rendered RGB values are then calculated as
\begin{equation}
   \label{eq:rendering}
   \hat{{\bf I}}_{i,c}({\bf X}, {\bf K}, {\bf L})=[K_{R}S({\bf N(\bf X),\bf L})L_R,K_GS({\bf N(\bf X),\bf L})L_G,K_BS({\bf N(\bf X),\bf L})L_B]^T,
\end{equation}
where $S$ is the shading calculated by using the second-order spherical harmonics illumination model~\cite{ramamoorthi2001efficient,wu2011high} as
\begin{equation}
   \label{eq:shading}
   \begin{aligned}
   S(\bf N(\bf X),\bf L)&=L_0+L_1N_y+L_2N_z+L_3N_x+L_4N_xN_y+L_5N_yN_z\\
   &+L_6(N_z^2-\frac{1}{3})+L_7N_xN_z+L_8(N_x^2-N_y^2), 
   \end{aligned}
\end{equation}
where ${\bf N}({\bf X}) = [N_x,N_y,N_z]^T$ represents the vertex's normal vector, which is calculated as the average of adjacent triangular patch's normals. Varying illuminations for each image and spatially varying albedos are considered as in~\cite{kim2016multi}.

\begin{figure}[t!]
\centering
\includegraphics[trim={0cm 0cm 0cm 0cm}, width=1\linewidth]{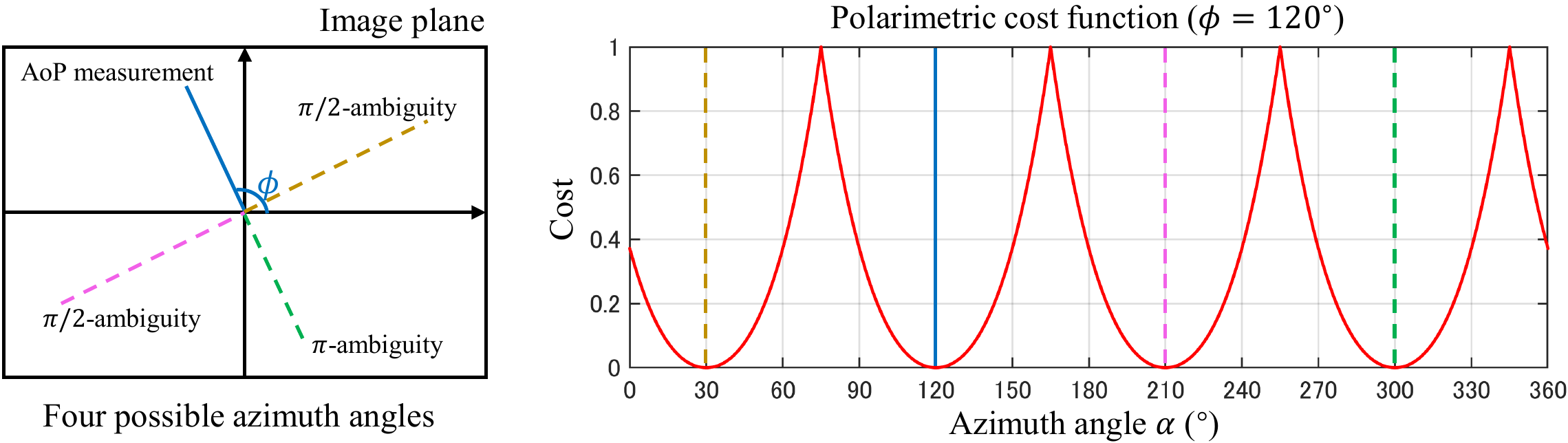}
\caption{An example of the polarimetric cost function
($\phi=120^{\circ}, k=0.5$). Four lines correspond to
possible azimuth angles as shown in Fig.~\ref{fig:ambiguity}.}
\label{fig:cost}
\end{figure}

\noindent
{\bf Polarimetric term:} To effectively constrain each estimated surface vertex's normal, we here propose a novel polarimetric term. Figure~\ref{fig:cost} shows an example of our polarimetric cost function for the case that the AoP measurement of the pixel corresponding to the vertex’s projection equals 120$^\circ$, i.e. $\phi=120^\circ$. This example corresponds to the situation as shown in Fig.~\ref{fig:ambiguity}. In both figures, four possible azimuth angles derived from the AoP measurement are shown by blue solid, purple dashed, green dashed, and brown dashed lines on the image plane, respectively. These four possibilities are caused by both the $\pi$-ambiguity and the $\pi/2$-ambiguity introduced in Section~\ref{sec:ambiguities}. In the ideal case without noise, one of the four possible azimuth angles should be the same as the azimuth angle of (unknown) true surface normal.

Based on this principle, as shown in Fig.~\ref{fig:cost}, our polarimetric term evaluates the difference between the azimuth angle of the estimated surface vertex's normal $\alpha$ and its closest possible azimuth angle from the AoP measurement~(i.e. $\phi-\pi/2,\ \phi,\ \phi+\pi/2$, or $\phi+\pi$). The cost function is mathematically defined as
\begin{equation}
   \label{eq:pol}
   E_{pol}({\bf X})=\sum_i \sum_{c\in \mathcal{V}(i)}
   \left(\frac{e^{-k\theta_{i,c}({\bf X})}-e^{-k}}{1-e^{-k}}\right)^2/{|\mathcal{V}(i)|},
\end{equation}
where $k$ is a parameter that determines the narrowness of the concave to assign the cost (see Fig.~\ref{fig:cost}). $\theta_{i,c}$ is defined as
\begin{equation}
   \label{eq:theta}
   \begin{aligned}
   \theta_{i,c}({\bf X})=1-4\eta_{i,c}({\bf X})/\pi,
   \end{aligned}
\end{equation}
where $\eta_{i,c}$ is expressed as
\begin{equation}
   \label{eq:eta}
   \begin{aligned}
   \eta_{i,c}({\bf X})=&\mathop{\min}(|\alpha_{i,c}({\bf N(\textbf X)})-\phi_{i,c}({\bf X})-\pi/2|,|\alpha_{i,c}({\bf N(\textbf X)})-\phi_{i,c}({\bf X})|,\\
   &|\alpha_{i,c}({\bf N(\textbf X)})-\phi_{i,c}({\bf X})+\pi/2|,|\alpha_{i,c}({\bf N(\textbf X)})-\phi_{i,c}({\bf X})+\pi|).
   \end{aligned}
\end{equation}
Here, $\alpha_{i,c}$ is the azimuth angle calculated by the projection of $i$-th vertex's normal to $c$-th image plane and $\phi_{i,c}$ is the corresponding AoP measurement.

Our polarimetric term mainly has two benefits. First, it enables us to constrain the estimated surface vertex's normal while simultaneously resolving the ambiguities based on the optimization using all vertices and all multi-view AoP measurements. Second, the concave shape of the cost function makes the normal constraint more robust to noise, which is an important property since AoP is susceptible to noise. The balance between the strength of the normal constraint and the robustness to noise can be adjusted by the parameter~$k$.

\noindent
{\bf Geometric smoothness term:}
The geometric smoothness term is applied to regularize the cost and to derive a smooth surface. This term is described as
\begin{equation}
   \label{eq:gsm}
   E_{gsm}({\bf X})=\sum_{m}
   \left(\frac{{\rm arccos}\left({\bf N}^\prime_m({\bf X})\cdot {\bf N}^\prime_{m_{avg}}({\bf X})\right)}{\pi}\right)^{q},
\end{equation}
where ${\bf N}^\prime_m$ represents the normal of $m$-th triangular patch, ${\bf N}^\prime_{m_{avg}}$ represents the averaged normal of its adjacent patches, and $q$ is a parameter to assign the cost. This term becomes small if the curvature of the surface is close to constant.

\noindent
{\bf Photometric smoothness term:}
Changes of pixel values in each image may result from different albedos or shading since spatially varying albedos are allowed in our model. To regularize this uncertainty, the same photometric smoothness term as \cite{kim2016multi} is applied as
\begin{equation}
   E_{psm}({\bf X,\bf K})=\sum_{i} \sum_{j\in \mathcal{A}(i)}
   w_{i,j}({\bf X})\left|\left|({\bf K}_i-{\bf K}_j)\right|\right|^2,
\end{equation}
where $\mathcal{A}(i)$ is the set of adjacent vertices of $i$-th vertex
and $w_{i,j}$ is the weight for the pair of $i$-th
and $j$-th vertices. A small weight is assigned, i.e. change of albedo is allowed, if a large chromaticity or intensity difference is observed between the corresponding pixels in the RGB image (see~\cite{kim2016multi} for details). By this term, a smooth variation in photometric information is considered as the result of shading while a sharp variation is considered as the result of varying albedos.

\section{Experimental Results}
\subsection{Implementation details}
We apply COLMAP~\cite{schonberger2016structure} for SfM and OpenMVS~\cite{OpenMVS} for MVS. The initial surface is reconstructed by the built-in surface reconstruction function of OpenMVS. The cost optimization of Eq.~(\ref{eq:costFunction}) is iterated three times by changing the weights as $(\tau_1, \tau_2, \tau_3)$ = $(0.05, 1.0, 1.0)$, $(0.1, 1.0, 1.0)$, and $(0.3, 1.0, 1.0)$. For each iteration, the parameter $q$ in Eq.~(\ref{eq:gsm}) is changed as $q$ = $2.2$, $2.8$, and $3.4$, while the parameter $k$ in Eq.~(\ref{eq:pol}) is set to 0.5 in all three iterations. By the three iterations, the surface normal constraint from AoP is gradually strengthened by allowing small normal variations to derive a fine shape while avoiding a local minimum. The non-linear optimization problem is solved by using Ceres solver~\cite{ceres-solver}.

\subsection{Comparison using synthetic data}

\begin{figure}[t!]
\centering
\includegraphics[trim={0cm 0cm 0cm 0cm}, width=1.0\linewidth]{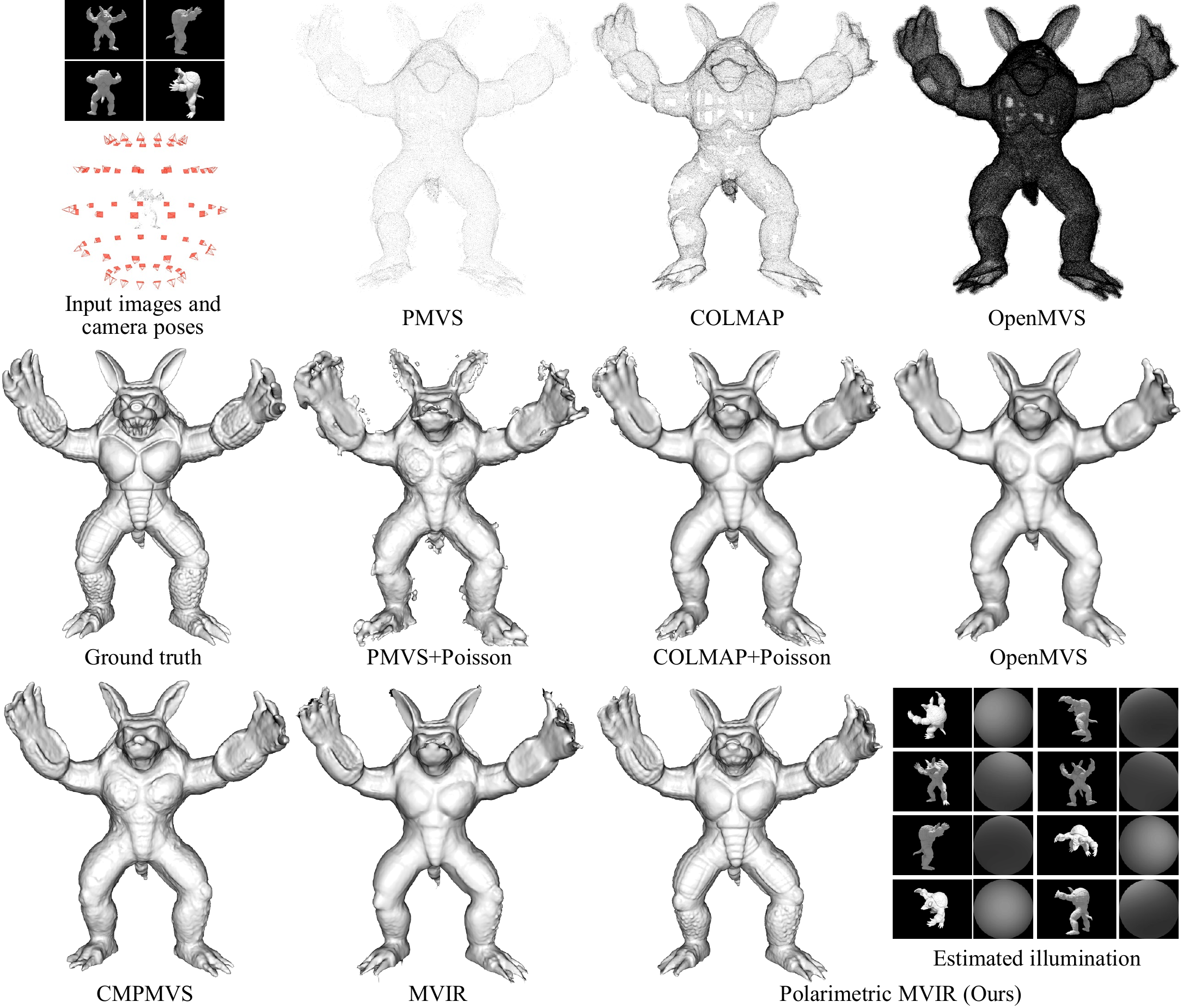}
\caption{Visual comparison for the Armadillo model}
\label{fig:evaluation}
\end{figure}

\setlength{\tabcolsep}{4pt}
\begin{table}[t!]
\begin{center}
\setlength\aboverulesep{0pt}\setlength\belowrulesep{0pt}
\setcellgapes{1.5pt}\makegapedcells
\caption{Comparisons of the average accuracy (Acc.) and completeness (Comp.) errors}
\label{table:evaluation}
\scalebox{0.78}{
\begin{tabular}{c|l!{\vrule width0.9pt}p{1.5cm}|p{1.5cm}|p{1.5cm}|p{1.5cm}|p{1.5cm}|p{1.5cm}}
\toprule[0.9pt]
\multicolumn{2}{l!{\vrule width0.9pt}}{} &
PMVS & CMPMVS & COLMAP & OpenMVS & MVIR & Ours\\
\midrule[0.9pt]
\midrule[0.9pt]
\multirow{3}{*}{Armadillo} & \# of Vertices & 60250 & 330733 & 268343 & 2045829 & 305555 & 305548\\
\cline{2-8}
& Acc.($\times10^{-2}$) & 1.2634 & 0.6287 & 0.7436 & 0.8503 & 0.7667 & {\textbf{0.4467}}\\
\cline{2-8}
& Comp.($\times10^{-2}$) & 1.5261 & 0.8676 & 1.0295 & 0.6893 & 0.9311 & \textbf{0.6365}\\
\midrule[0.9pt]
\multirow{3}{*}{Bunny} 
& \# of Vertices & 92701 & 513426 & 334666 & 2394638 & 399864 & 399863\\
\cline{2-8}
& Acc.($\times10^{-2}$) & 1.0136 & 0.7766 & 0.7734 & 1.0222 & 0.7629 & {\textbf{0.5706}}\\
\cline{2-8}
& Comp.($\times10^{-2}$) & 1.3873 & 0.9581 & 1.6987 & 0.8466 & 0.8118 & \textbf{0.6447}\\
\midrule[0.9pt]
\multirow{3}{*}{Dragon} & \# of Vertices & 88519 & 474219 & 399624 & 2820589 & 460888 & 460667\\
\cline{2-8}
& Acc.($\times10^{-2}$) & 1.4321 & 0.8826 & 0.9001 & 1.0421 & 0.8563 & {\textbf{0.6258}}\\
\cline{2-8}
& Comp.($\times10^{-2}$) & 2.0740 & 1.4036 & 1.6606 & 1.3179 & 1.2237 & \textbf{1.0222}\\
\midrule[0.9pt]
\multirow{3}{*}{Buddha} & \# of Vertices & 61259 & 338654 & 320539 & 2204122 & 348967 & 348691\\
\cline{2-8}
& Acc.($\times10^{-2}$) & 1.7658 & 1.0565 & 0.9658 & 1.0878 & 1.0588 & {\textbf{0.7926}}\\
\cline{2-8}
& Comp.($\times10^{-2}$) & 2.4254 & 1.5666 & 2.0094 & 1.4859 & 1.3968 & \textbf{1.1487}\\
\midrule[0.9pt]
\midrule[0.9pt]
\multirow{2}{*}{Average} 
& Acc.($\times10^{-2}$) & 1.3687 & 0.8361 & 0.8457 & 1.0006 & 0.8612 & {\textbf{0.6089}}\\
\cline{2-8}
& Comp.($\times10^{-2}$) & 1.8532 & 1.1990 & 1.5996 & 1.0849 & 1.0909 & \textbf{0.8630}\\
\bottomrule[0.9pt]
\end{tabular}}
\end{center}
\end{table}
\setlength{\tabcolsep}{1.4pt}

\setlength{\tabcolsep}{4pt}
\begin{table}[t!]
\begin{center}
\setlength\aboverulesep{0pt}\setlength\belowrulesep{0pt}
\setcellgapes{1pt}\makegapedcells
\caption{Numerical evaluation with 50\% ambiguity and Gaussian noise}
\label{table:robust}
\scalebox{0.78}{
\begin{tabular}{p{8cm}!{\vrule width0.9pt}p{2.2cm}|p{2.2cm}}
\toprule[0.9pt]
 & Acc.($\times10^{-2}$) & Comp.($\times10^{-2}$)\\
\midrule[0.9pt]
\midrule[0.9pt]
CMPMVS (Best accuracy in the existing methods) & 0.8361 & 1.1990 \\
OpenMVS (Best completeness in the existing methods) & 1.0006 & 1.0849\\
Ours (No ambiguity $\&$ $\sigma = 0^\circ$) & \textbf{0.6089} & \textbf{0.8630}\\
\midrule[0.9pt]
Ours ($50\%$ ambiguity $\&$ $\sigma=0^\circ$) & 0.6187 & 0.8727 \\
Ours ($50\%$ ambiguity $\&$ $\sigma=6^\circ$) & 0.6267 & 0.8820\\
Ours ($50\%$ ambiguity $\&$ $\sigma=12^\circ$) & 0.6367 & 0.8892\\
Ours ($50\%$ ambiguity $\&$ $\sigma=18^\circ$) & 0.6590 & 0.9115\\
Ours ($50\%$ ambiguity $\&$ $\sigma=24^\circ$) & 0.7175 & 0.9643\\
\bottomrule[0.9pt]
\end{tabular}}
\end{center}
\end{table}
\setlength{\tabcolsep}{1.4pt}

Numerical evaluation was performed using four CG models (Armadillo, Stanford bunny, Dragon, and Buddha) available from Stanford 3D Scanning Repository~\cite{Stanford}. 
Original 3D models were subdivided to provide enough number of vertices as ground truth. Since it is very difficult to simulate realistic polarization images and there are no public tools and datasets for polarimetric 3D reconstruction, we synthesized the RGB and the AoP inputs using Blender~\cite{Blender} as follows. 
Using spherically placed cameras, the RGB images were rendered under a point light source located at infinity and an environmental light uniformly contributing to the surface (see Fig.~\ref{fig:evaluation}). 
For synthesizing AoP images, AoP for each pixel was obtained from the corresponding azimuth angle, meaning that there is no $\pi/2$-ambiguity, for the first experiment. The experiment was also conducted by randomly adding ambiguities and Gaussian noise to the azimuth angles.

We compared our Polarimetric MVIR with four representative MVS methods (PMVS~\cite{furukawa2009accurate}, CMPMVS~\cite{jancosek2011multi}, MVS in COLMAP~\cite{schonberger2016pixelwise}, OpenMVS~\cite{OpenMVS}) and MVIR~\cite{kim2016multi} using the same initial model as ours. Ground-truth camera poses are used to avoid the alignment problem among the models reconstructed from different methods. Commonly used metrics~\cite{aanaes2016large,ley2016syb3r},
i.e. accuracy which is the distance from each estimated 3D point to its nearest ground-truth 3D point and completeness which is the distance from each ground-truth 3D point to its nearest estimated 3D point, were used for evaluation. As estimated 3D points, the output point cloud was used for PMVS, COLMAP, and OpenMVS, while the output surface's vertices were used for CMPMVS, MVIR and our method.

Table~\ref{table:evaluation} shows the comparison of the average accuracy and the average completeness for each model. The results show that our method achieves the best accuracy and completeness for all four models with significant improvements. Visual comparison for Armadillo is shown in Fig.~\ref{fig:evaluation}, where the surfaces for PMVS and COLMAP were created using Poisson surface reconstruction~\cite{kazhdan2013screened} with our best parameter choice, while the surface for OpenMVS was obtained using its built-in function.  We can clearly see that our method can recover more details than the other methods by exploiting AoP information. The visual comparison for the other models can be seen in our supplementary material.

Table~\ref{table:robust} shows the numerical evaluation for our method when 50\% random ambiguities and Gaussian noise with different noise levels were added to the azimuth-angle images. Note that top three rows are the results without any disturbance and same as those in Table 1, while the bottom five rows are the results of our method with ambiguity and noise added on AoP images. These results demonstrate that our method is quite robust against the ambiguity and the noise and outperforms the best-performed existing methods
even with the 50\% ambiguities and a large noise level ($\sigma = 24^\circ$).

\subsection{Comparison using real data}
\begin{figure}[t!]
\centering
\includegraphics[trim={0cm 0cm 0cm 0cm}, width=1\linewidth]{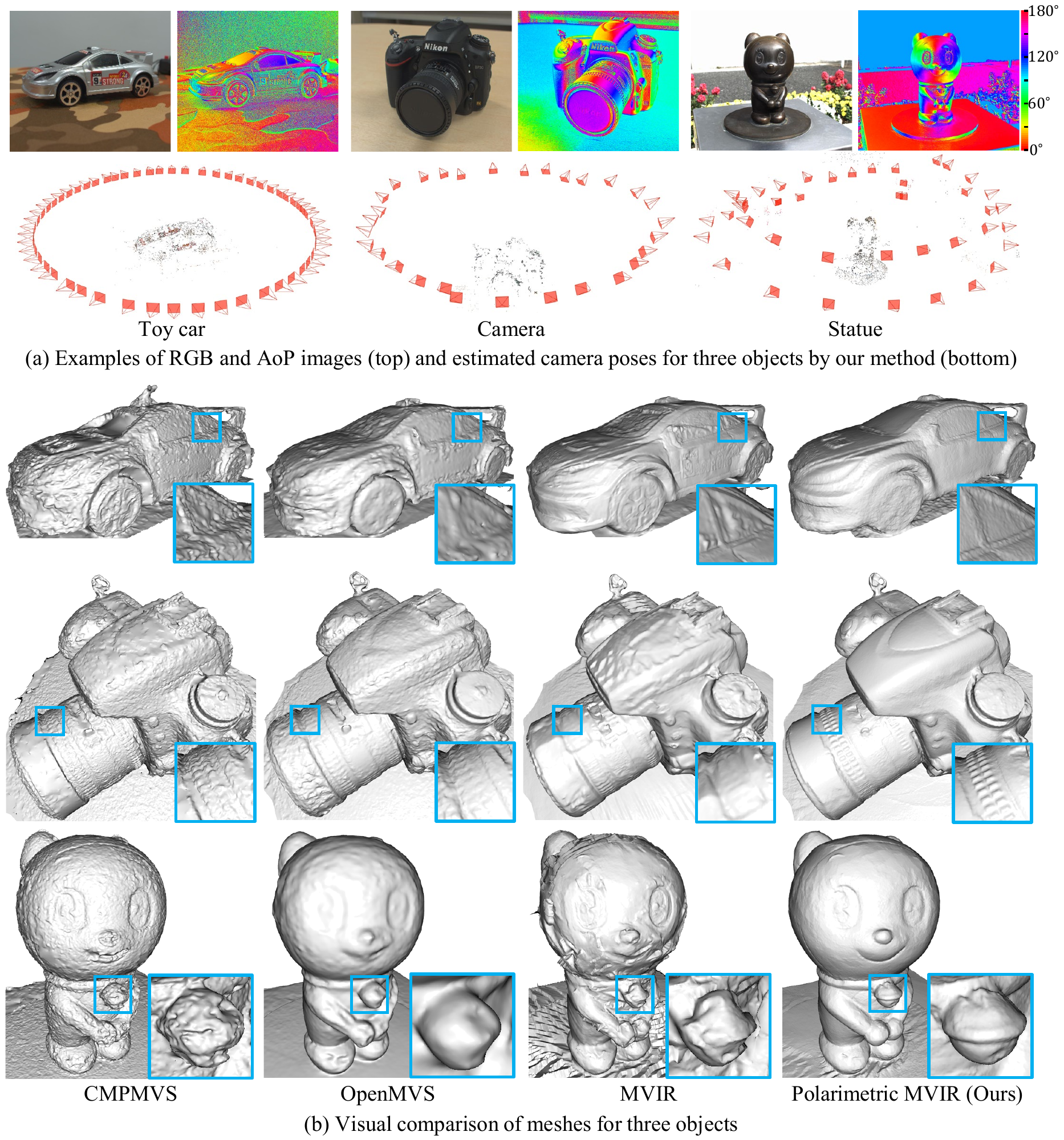}
\caption{Visual comparison using real data for the three objects}
\label{fig:result}
\end{figure}

Figure~\ref{fig:result} shows the visual comparison of the reconstructed 3D models using real images of a toy car (56 views) and a camera (31 views) captured under a normal lighting condition in the office using fluorescent light on the ceiling, and a statue (43 views) captured under outdoor daylight with cloudy weather.
We captured the polarization images using Lucid PHX050S-Q camera~\cite{Lucid}. We compared our method with CMPMVS, OpenMVS, and MVIR, which respectively provide the best accuracy, the best completeness, and the best balanced result among the existing methods shown in Table~\ref{table:evaluation}. The results of all compared methods and our albedo and illumination results can be seen in the supplementary material. 

The results of Fig.~\ref{fig:result} show that CMPMVS can reconstruct fine details in relatively well-textured regions (e.g. the details of the camera lens), while it fails in texture-less regions (e.g. the front window of the car). OpenMVS can better reconstruct the overall shapes owing to the denser points, although some fine details are lost. MVIR performs well except for dark regions, where the shading information is limited (e.g. the top of the camera and the surface of the statue). On the contrary, our method can recover finer details and clearly improve the reconstructed 3D model quality by exploiting both photometric and polarimetric information, especially in regions such as the front body and the window of the toy car, and the overall surfaces of the camera and the statue.

\begin{figure}[t!]
\centering
\includegraphics[trim={0cm 0cm 0cm 0cm}, angle=0, width=0.98\linewidth]{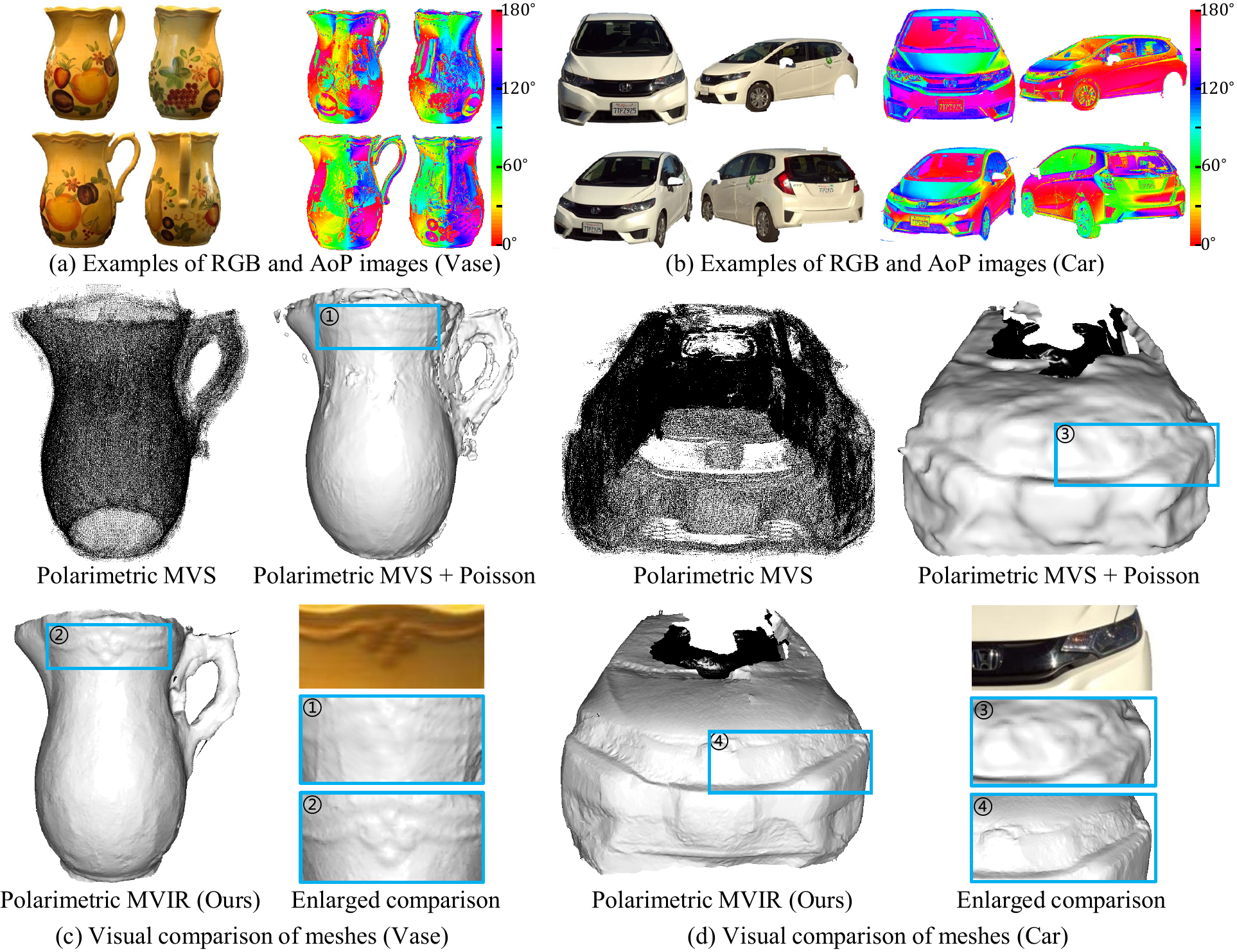}
\caption{Comparison with Polarimetric MVS~\cite{cui2017polarimetric} using the data provided by the authors.}
\label{fig:pmvs}
\end{figure}

\subsection{Refinement for Polarimetric MVS~\cite{cui2017polarimetric}}
Since Polarimetric MVS~\cite{cui2017polarimetric} can be used for our initial model to make better use of polarimetric information,
we used point cloud results of two objects (vase and car) obtained by Polarimetric MVS for the initial surface generation and then refined the initial surface using the provided camera poses, and RGB and AoP images from 36 viewpoints as shown in Fig.~\ref{fig:pmvs} (a) and (b).
As shown in Fig.~\ref{fig:pmvs} (c) and (d),
Polarimetric MVS can provide dense point clouds, even for texture-less regions, by exploiting polarimetric information. However, there are still some outliers, which could be derived from AoP noise and incorrect disambiguation, and resultant surfaces are rippling. These artifacts are alleviated in our method (Polarimetric MVIR) by solving the ambiguity problem in our global optimization. Moreover, we can see that finer details are reconstructed using photometric shading information in our cost function.

\section{Conclusions}

In this paper, we have proposed Polarimetric MVIR, which can reconstruct a high-quality 3D model by optimizing multi-view photometric rendering errors and polarimetric errors. Polarimetric MVIR resolves the $\pi$- and $\pi/2$-ambiguities as an optimization problem, which makes the method fully passive and applicable to various materials. Experimental results have demonstrated that Polarimetric MVIR is robust to ambiguities and noise, and generates more detailed 3D models compared with existing state-of-the-art multi-view reconstruction methods.

Our Polarimetric MVIR has a limitation that it requires a reasonably good initial shape for its global optimization, which  would encourage us to develop more robust initial shape estimation.

\noindent \\
\textbf{Acknowledgment}
This work was partly supported by JSPS KAKENHI Grant Number 17H00744. The authors would like to thank Dr. Zhaopeng Cui for sharing the data of Polarimetric MVS.

\clearpage

\bibliographystyle{splncs04}
\bibliography{egbib}
\end{document}